\documentclass{article}
\PassOptionsToPackage{numbers, compress}{natbib}

\usepackage[preprint]{neurips_2023}

\usepackage[utf8]{inputenc} 
\usepackage[T1]{fontenc}    
\usepackage{url}            
\usepackage{booktabs}       
\usepackage{amsfonts}       
\usepackage{nicefrac}       
\usepackage{microtype}      
\usepackage{xcolor}         

\usepackage{amsmath}
\usepackage{amssymb}
\usepackage{graphicx}
\usepackage{multirow}
\usepackage{colortbl}
\usepackage{makecell}
\usepackage{enumitem}
\usepackage{hyperref} 
\usepackage{caption}
\hypersetup{
hidelinks,
colorlinks=true,
linkcolor=red,
citecolor=green
}

\title{HG$^{3}$-NeRF: Hierarchical Geometric, Semantic, and Photometric Guided Neural Radiance Fields \\ for Sparse View Inputs}

\author{
Zelin Gao$^{1}$,
~Weichen Dai$^{2}$,
~Yu Zhang$^{1}$\thanks{Corresponding authors.}
\vspace{2pt}\\
$^{1}$ College of Control Science and Engineering, Zhejiang University\\
$^{2}$ School of Computer Science, Hangzhou Dianzi University
}

\begin{document}

\maketitle

\begin{abstract}
Neural Radiance Fields~(NeRF) have garnered considerable attention as a paradigm for novel view synthesis by learning scene representations from discrete observations. Nevertheless, NeRF exhibit pronounced performance degradation when confronted with sparse view inputs, consequently curtailing its further applicability. In this work, we introduce \textbf{H}ierarchical \textbf{Geometric}, \textbf{Semantic}, and \textbf{Photometric} \textbf{G}uided \textbf{NeRF}~(HG$^{3}$-NeRF), a novel methodology that can address the aforementioned limitation and enhance consistency of geometry, semantic content, and appearance across different views. We propose Hierarchical Geometric Guidance~(HGG) to incorporate the attachment of Structure from Motion~(SfM), namely sparse depth prior, into the scene representations. Different from direct depth supervision, HGG samples volume points from local-to-global geometric regions, mitigating the misalignment caused by inherent bias in the depth prior. Furthermore, we draw inspiration from notable variations in semantic consistency observed across images of different resolutions and propose Hierarchical Semantic Guidance~(HSG) to learn the coarse-to-fine semantic content, which corresponds to the coarse-to-fine scene representations. Experimental results demonstrate that HG$^{3}$-NeRF can outperform other state-of-the-art methods on different standard benchmarks and achieve high-fidelity synthesis results for sparse view inputs.
\end{abstract}

\section{Introduction}
Novel View Synthesis~(NVS) is one of the crucial tasks in computer vision, aiming to generate images of unseen views through visual information, similar to how humans perceive and visualize their surroundings. Recent methods have opted to recover intermediate dense 3D-aware representation~\cite{xu20223d,lee2022exp,or2022stylesdf}, multi-plane images~\cite{xing2022temporal,solovev2023self,han2022single}, or volume density~\cite{barron2022mip,muller2022instant,mildenhall2020nerf}, followed by neural rendering theorem~\cite{peng2022bokehme,suhail2022light,eslami2018neural} to synthesize images. In particular, Neural Radiance Fields~(NeRF)~\cite{mildenhall2020nerf} have demonstrated remarkable potential with state-of-the-art performance in generating high-fidelity view synthesis results. However, NeRF suffer from significant challenges with sparse view inputs, primarily due to their reliance on dense scene coverage to mitigate the shape ambiguity problem~\cite{wei2021nerfingmvs,zhang2020nerf++} arising from only photometric supervision.

\begin{figure}[!ht]
\centering
\includegraphics[width=1.\linewidth]{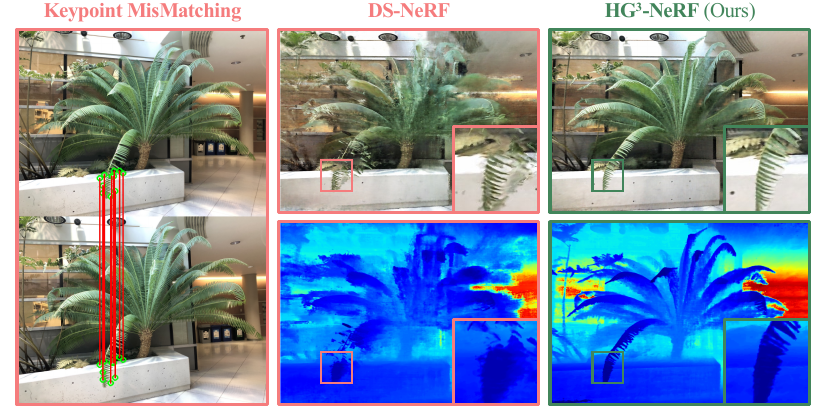}
\caption{\textbf{Novel View Synthesis Results from 3 View Inputs.}
Left: The bias in the sparse depth prior is caused by keypoint mismatching during the multi-view stereo process of SfM. Mid: Bias in the depth prior is introduced to NeRF through depth supervision and leads to geometric misalignment. Right: Our HG$^{3}$-NeRF leverage additional guidance to learn the scene representations, showing great performance in both appearance and geometry.}
\label{fig:intro}
\vspace{-10pt}
\end{figure}

Existing methods address this challenging issue with a variety of strategies, which can be classified into pre-training methods~\cite{liu2022neural,johari2022geonerf,chen2021mvsnerf,chibane2021stereo,yu2021pixelnerf} and per-scene optimization methods~\cite{niemeyer2022regnerf,kim2022infonerf,jain2021putting}. Pre-training methods train the model on large-scale datasets and further fine-tune it for each scene at test time. However, the generalization ability heavily depends on the quality of datasets, and it is too expensive to obtain necessary datasets by capturing many different scenes. The alternative methods are to train the network from scratch for each scene. Though most of these methods introduce additional regularization to prevent overfitting issues, they still suffer from geometric misalignment without real-world geometric supervision. Therefore, some methods~\cite{deng2022depth} leverage the supervision function contributed by the rough depth prior estimated by Structure from Motion~(SfM)~\cite{schonberger2016structure,agarwal2011building} to improve the geometric consistency by enforcing the distribution of density and color around the depth prior. However, as shown in the left and mid of Fig.~\ref{fig:intro}, direct depth supervision can also introduce bias generated from the multi-view stereo process~\cite{mur2017orb,mur2015orb,hartley2003multiple} into the scene representations, resulting in geometric misalignment and poor view synthesis quality.

It should also be noted that NeRF is actually able to recover reasonable geometric results with dense view inputs~\cite{wei2021nerfingmvs,chen2021mvsnerf}. However, given the sparse viewpoints, it is too difficult for NeRF to learn the consistent color and density distribution across different views with only photometric supervision. Due to the fact that the sparse depth prior can indicate the approximate distribution of the color and density along the ray, the sparse depth prior should be used to guide the volume sampling to improve the geometric consistency. As shown in the right of Fig.~\ref{fig:intro}, sampling volume points with the guidance of the sparse depth prior is a more precise strategy to improve geometric consistency rather than direct depth supervision.

In this paper, we exploit the geometric, semantic, and photometric guidance to represent the neural radiance fields from sparse view inputs. We propose hierarchical geometric guidance~(HGG) to sample volume points with the depth prior, which is generated as a common attachment in NeRF pipelines by running SfM. Since the depth prior provides approximate locations of the density and color along the rays, we first sample points around the depth prior in a local region to initialize the neural radiance fields and then gradually extend the sampling region to full scene bounds for learning the global scene representations. Different from direct depth supervision, the HGG method utilizes a local-to-global sampling strategy to incorporate the depth prior into the representations and mitigate the potential geometric misalignment caused by bias in the depth prior. Furthermore, we propose hierarchical semantic guidance~(HSG) to supervise semantic consistency of the complex real-world scenarios using CLIP~\cite{radford2021learning}. We draw inspiration from the diverse variations in semantic consistency observed across images of different resolutions, where the semantic content of low-resolution images is hard to match with that of high-resolution images. Since the scene representations are learned from coarse to fine, the rendered images are blurred like low-resolution images at the start of training~\cite{lin2021barf}. Therefore, the HSG method first leverages coarse feature vectors from the down-sampled images to supervise the semantic consistency. As the iteration increases, it gradually aggregates more content into the feature vectors by reducing the down sampling rate. Finally, we adopt the hierarchical photometric guidance proposed in NeRF to supervise the appearance consistency. Combined, we call our method HG$^{3}$-NeRF, which incorporate hierarchical geometric, semantic, and photometric guidance to represent the neural radiance fields from sparse view inputs. 

In the experiments, we evaluate the effectiveness of the proposed HG$^{3}$-NeRF in comparison to state-of-the-art baselines on various standard benchmarks. Furthermore, we conduct the model analysis including a comprehensive ablation study to investigate the contributions of HGG and HSG, respectively, as well as a view synthesis comparison study to demonstrate that our HGG method can achieve comparable performance to represent forward-facing scenarios in real-world space instead of using NDC~(Normalized Device Coordinate) space.

To summarize, our contributions are as follows:
\begin{itemize}[leftmargin=*]
    \item We propose HG$^{3}$-NeRF, a novel methodology that can exploit the hierarchical geometric, semantic, and photometric guidance to maintain consistency across different views to represent the neural radiance fields from sparse view inputs.
    \item We propose hierarchical geometric guidance~(HGG) that incorporates sparse depth prior to the scene representations without introducing bias and hierarchical semantic guidance~(HSG) that guides the coarse-to-fine semantic supervision of complex real-world scenarios.
    \item We conduct the experiments on various datasets and show that the proposed HG$^{3}$-NeRF can achieve realistic synthesis results for sparse viewpoint inputs.
\end{itemize}

\section{Related Works}
\textbf{Neural Scene Representations.}
In computer vision, coordinate-based neural representations~\cite{sitzmann2020implicit,chen2019learning,mescheder2019occupancy,michalkiewicz2019implicit} have become one of the most popular representation methods for various 3D vision tasks, such as 3D reconstruction~\cite{yu2022monosdf,sun2022direct,wang2021neus}, 3D-aware generation~\cite{xu20223d,lee2022exp,or2022stylesdf} and novel view synthesis~\cite{barron2022mip,muller2022instant,niemeyer2022regnerf,mildenhall2020nerf}. As opposed to explicit representations such as point clouds~\cite{peng2021shape,groueix2018papier,fan2017point}, voxels~\cite{choy20163d,wu20153d} or meshes~\cite{deng2020cvxnet,nash2020polygen,bagautdinov2018modeling}, this paradigm signifies that color and 3D geometric information can be represented by the implicit neural network, leading to a more compact representation format. There are several works~\cite{liu2019learning,mildenhall2020nerf,niemeyer2020differentiable,sitzmann2019deepvoxels} learning the scene representations from multi-view images using neural volume rendering. Among these methods, Neural Radiance Fields~(NeRF) have demonstrated remarkable potential in generating high-fidelity images from novel viewpoints as well as its simplicity to represent the scene as a continuous implicit function. Therefore, various follow-up works have been proposed to improve the performance and generality of NeRF such as large-scale scene representations from city or satellite viewpoints~\cite{tancik2022block,xiangli2022bungeenerf}, real-time neural volume rendering for fast training~\cite{muller2022instant,fridovich2022plenoxels}, and unbounded scene representations~\cite{barron2022mip,zhang2020nerf++}. Although these methods have extended NeRF to various domains with impressive performance, they typically require dense viewpoints to learn accurate scene representations for synthesizing realistic images, limiting their further application in real-world scenarios~\cite{niemeyer2022regnerf,kim2022infonerf}. In this work, we focus on solving the issue where only sparse view inputs are available for NeRF, which is much closer to real-world applications.

\textbf{Novel View Synthesis from Sparse View Inputs.}
To tackle the problem of representing the neural radiance fields from sparse view inputs, several works are proposed for the further real-world application of NeRF, which can be classified into two main categories: pre-training methods and per-scene optimization methods. The pre-training methods~\cite{liu2022neural,johari2022geonerf,jang2021codenerf,rematas2021sharf,chen2021mvsnerf,chibane2021stereo,yu2021pixelnerf} circumvent the requirement of dense view inputs by pre-training a conditional model to aggregate sufficient prior knowledge for reconstructing the neural radiance fields. They propose to train a generalizable model with the high-dimensional image features extracted from a CNN backbone network~\cite{chibane2021stereo,yu2021pixelnerf} or the 3D cost volume obtained by image warping~\cite{chen2021mvsnerf,johari2022geonerf,liu2022neural}. Though these methods achieve great results for sparse view inputs, large-scale datasets of many different scenarios are required for pre-training. Moreover, these methods require fine-tuning the network parameters and suffer from quality degradation on untrained domains. The per-scene optimization methods~\cite{seo2023mixnerf,niemeyer2022regnerf,kim2022infonerf,xu2022sinnerf,jain2021putting} propose to supervise NeRF with additional regularization loss function instead of using expensive pre-training models. Although various regularization functions over geometry~\cite{xu2022sinnerf,niemeyer2022regnerf}, appearance~\cite{niemeyer2022regnerf}, semantics~\cite{xu2022sinnerf,jain2021putting}, and density distribution~\cite{kim2022infonerf} are used to supervise the consistency between seen and unseen viewpoints, the performance on geometric consistency is still hard to be improved due to the lack of real-world geometric supervision. Some methods~\cite{deng2022depth} directly supervise the geometry of NeRF with the depth prior estimated by SfM, however, geometric misalignment still exists since the bias in the depth prior is introduced into the scene representations during the optimization.

\begin{figure}[!t]
\centering
\includegraphics[width=1.\linewidth]{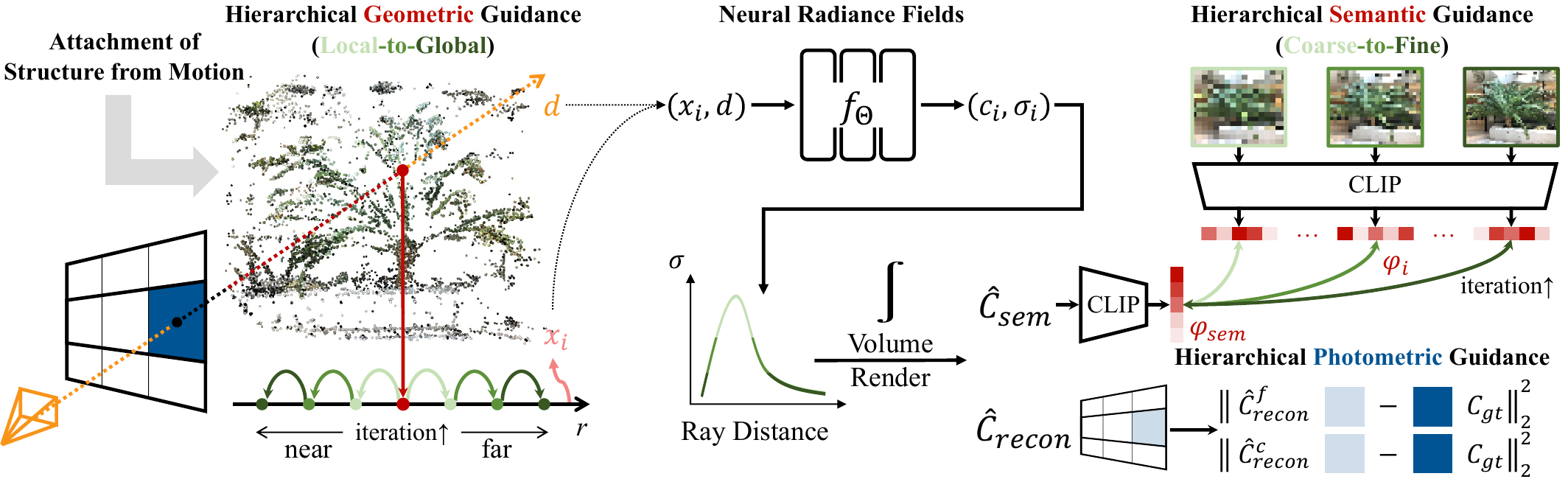}
\caption{\textbf{Overview of HG$^{3}$-NeRF.} 
The 3D volume location $\mathbf{x}_i$ is first sampled within the local-to-global region setup by hierarchical geometric guidance and then fed into neural radiance fields along with the viewing direction $\mathbf{d}$ to query color $\mathbf{c}_i$ and density $\sigma_{i}$. Via the volume rendering theorem, the query results are integrated into $\mathbf{\hat{C}}_{recon}$, which contains $\mathbf{\hat{C}}_{recon}^{c}$ for the coarse model and $\mathbf{\hat{C}}_{recon}^{f}$ for the fine model. Moreover, we employ CLIP to encode the image $\mathbf{\hat{C}}_{sem}$ rendered from a randomly selected pose as a feature vector $\varphi_{sem}$. The scene representations are finally optimized by the coarse-to-fine cosine similarity between $\varphi_{sem}$ and $\varphi_{i}$ from hierarchical semantic guidance as well as the MSE between $\mathbf{\hat{C}}_{recon}$ and the observed color $\mathbf{C}_{gt}$.}
\label{fig:over}
\vspace{-10pt}
\end{figure}

\section{Preliminaries}
\subsection{Neural Radiance Fields} 
In this work, we follow the framework proposed in NeRF to integrate the predicted color and density along the ray. Specifically, NeRF adopt multi-layer perceptrons~(MLPs) to reconstruct a radiance field from dense input views, where the view-dependent appearance is modeled as a continuous function $f_{\Theta}: (\mathbf{x}_{i}, \mathbf{d}) \xrightarrow{}(\mathbf{c}_{i},\sigma_{i})$, which maps a 3D location $\mathbf{x}_{i}=(x_i, y_i, z_i)$ with its 2D viewing direction $\mathbf{d}=(\theta, \phi)$ into a volume color $\mathbf{c}_{i}=(r_i,g_i,b_i)$ and a density $\sigma_{i}$. 

\textbf{Neural Volume Rendering.}
Given a ray $\mathbf{r}(t)=\mathbf{o}+t\mathbf{d}$ whose origin is at the camera’s center of projection $\mathbf{o}$, a volume location $t_{i} \in [t_{n}, t_{f}]$ is sampled within the near and far planes. By querying $f_{\Theta}$, the final color $\mathbf{\hat{C}}_{recon}(\mathbf{r})$ is approximated via the volume rendering theorem as
\begin{equation}
    \mathbf{\hat{C}}_{recon}(\mathbf{r})
    =
    \sum_{i=1}^{N} T_{i}\left(1-\exp \left(-\sigma_{i} \delta_{i}\right)\right) \mathbf{c}_{i}
    \text{, where }
    T_{i} = \exp \left(-\sum_{j=1}^{i-1} \sigma_{j} \delta_{j}\right)
\label{eq:vr}
\end{equation}
Note that $T_{i}$ denotes how much light is transmitted on ray $\mathbf{r}$ up to sample $i$ and $\delta_{i}=t_{i+1}-t_{i}$ denotes the interval between the $i$-th sample and its adjacent one.

\textbf{Hierarchical Photometric Guidance.} To improve the performance of NVS, NeRF propose to train a coarse model and a fine model, with parameters $\Theta^{c}$ and $\Theta^{f}$, respectively. The hierarchical photometric guidance~(HPG) can be formulated as:
\begin{equation}
\mathcal{L}_{hpg}
=
\sum_{\mathbf{r} \in \mathcal{R}} 
\left(\left\|\mathbf{C}_{gt}(\mathbf{r})-\mathbf{\hat{C}}^{c}_{recon}\left(\mathbf{r};\mathbf{t}^c\right)\right\|_2^2
+ 
\left\|\mathbf{C}_{gt}(\mathbf{r})-\mathbf{\hat{C}}^{f}_{recon}\left(\mathbf{r};\mathbf{t}^c \cup \mathbf{t}^f\right)\right\|_2^2\right)
\label{eq:hpg}
\end{equation}
where $\mathcal{R}$ denotes the set of all rays across all images, $\mathbf{\hat{C}}^{c}_{recon}(\mathbf{r};\mathbf{t}^c)$ denotes the predicted color of the coarse model with stratified volume samples $\mathbf{t}^c$, and $\mathbf{\hat{C}}^{f}_{recon}(\mathbf{r};\mathbf{t}^c\cup\mathbf{t}^f)$ denotes the predicted color of the fine model with the union of $\mathbf{t}^c$ and $\mathbf{t}^f$ which is produced by inverse transform sampling.

\subsection{Contrastive Language-Image Pre-Training} 
Contrastive Language-Image Pre-Training~(CLIP) model is trained for learning joint representations between text and images~\cite{nichol2021glide}. CLIP consist of an image encoder and a text encoder, demonstrating the excellent few-shot transfer performance to image recognition tasks~(for brevity, the following CLIP only refers to its image encoder $f_{clip}$). Though the semantic consistency implemented by CLIP can improve the performance of object-level scene representations~\cite{jain2021putting,xu2022sinnerf}, it is still severely limited for complex real-world scenarios~\cite{niemeyer2022regnerf}. 

\section{Methodology}
In this paper, we propose HG$^{3}$-NeRF to represent the neural radiance fields for sparse view inputs. The overview of our method is illustrated in Fig.~\ref{fig:over}. Given the attachment of SfM, i.e., sparse depth prior, and a ray generated from the camera pose, we first sample volume points within the local-to-global region using hierarchical geometric guidance~(Sec.~\ref{sec:HGG}) and then query the corresponding results to predict the final color via the volume rendering theorem. We also randomly select a camera pose to render an image and encode it into the corresponding feature vector using CLIP $f_{clip}$. Finally, we compute the coarse-to-fine semantic cosine similarity under hierarchical semantic guidance~(Sec.~\ref{sec:HSG}) and the appearance MSE under hierarchical photometric guidance~(Eq.~\ref{eq:hpg}) to optimize the neural radiance fields.

\subsection{Hierarchical Geometric Guidance} \label{sec:HGG}
We propose hierarchical geometric guidance~(HGG), which incorporates the geometric consistency into the scene representations by using sparse depth prior from SfM. The HGG method guides NeRF to learn the approximate distribution of the density and color from a local-to-global sampling region determined by the depth prior.

\begin{figure}[!t]
\centering
\includegraphics[width=.9\linewidth]{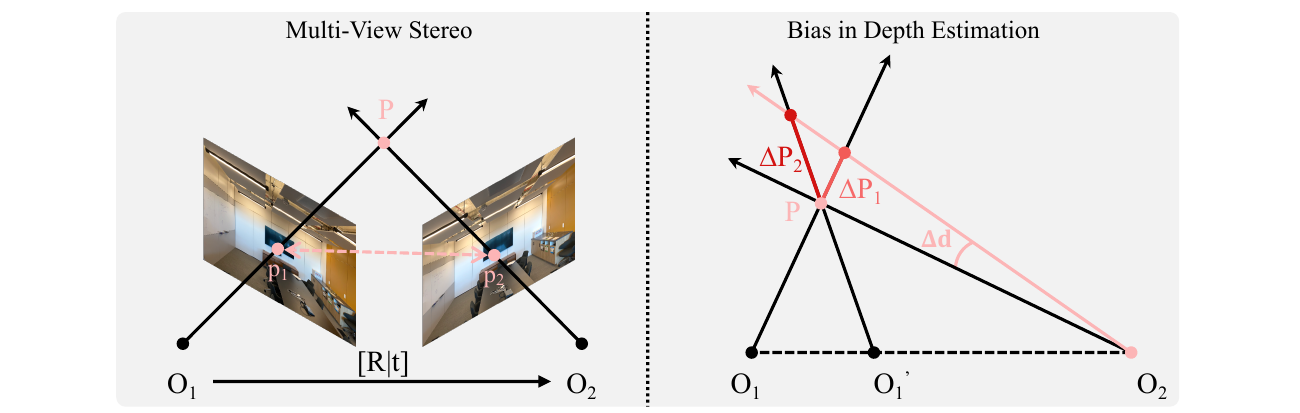}
\caption{\textbf{Bias from Multi-View Stereo.} Classical stereo methods estimate depth through geometric constraints based on keypoint matching. The shift $\Delta{d}$ in ray direction is caused by keypoint mismatching and thus introduces the bias into the depth estimation. Especially when the translation between two frames is small, the bias further increases.}
\label{fig:hgg}
\vspace{-10pt}
\end{figure}

As the common attachment of SfM, the sparse depth prior is estimated by the multi-view triangulation method, which can be formulated as:
\begin{equation}
    s_{1} p_{1} = s_{2}Rp_{2} + t
\end{equation}
where $p_1,p_2$ denote the locations of two matched keypoints on the normalized coordinate, $s_1,s_2$ denote the depth values, and $R,t$ denote rotation matrix and translation vector between these two points. As shown in Fig.~\ref{fig:hgg}, the accuracy of estimated 3D point $P$ relies on the quality of keypoint matching. The keypoint mismatching can causes a shift $\Delta{d}$ in the ray direction and thus introduces a bias $\Delta{P_1}$ into the estimated depth. In addition, the small translation between two frames can further increase the bias to $\Delta{P_2}$. 

The depth estimation of NeRF can be formulated as:
\begin{equation}
    \mathbf{\hat{D}}_{recon}(\mathbf{r})
    =
    \sum_{i=1}^{N} T_{i}\left(1-\exp \left(-\sigma_{i} \delta_{i}\right)\right) t_{i} 
\end{equation}
Although sparse depth prior from SfM can be used to supervise $\mathbf{\hat{D}}_{recon}$, the optimization function contributed by this sparse depth prior can introduce the aforementioned bias into the scene representations, resulting in geometric misalignment and poor view synthesis results.

To address this issue, the HGG method utilizes sparse depth prior to indicate the sampling region for learning the density distribution along the ray instead of direct supervision. 
We first initialize the scene representations by setting near planes $t_{n}$ and far planes $t_{f}$ close to the depth prior for sampling volume points from a local region. Then, $t_{n}$ and $t_{f}$ are gradually extended to the full scene bounds for learning the global volume distribution along the ray, which can be formulated as:
\begin{equation}
\begin{aligned}
t_{n}(i) & =t_{depth}+\left(t_{n}-t_{depth}\right) \gamma(i) \\
t_{f}(i) & =t_{depth}+\left(t_{f}-t_{depth}\right) \gamma(i)
\end{aligned}
\end{equation}
where $\gamma(i)$ denotes the region adjustment rate at $i$ iteration, which can be formulated as:
\begin{equation}
\gamma(i) 
=
\frac{1-\cos\left(\left({\min \left( \max \left( i/N_{hgg},\epsilon_{hgg} \right),1 \right)}\right)\pi\right)}{2}
\end{equation}
where $\epsilon_{hgg}$ denotes the minimum adjusting rate, and $N_{hgg}$ denotes how many iterations until the full bounds are reached. The HGG method can utilize the information provided by the depth prior, i.e., the approximate density distribution of the rays, and reconstruct the neural radiance fields with the local-to-global geometric guidance of the real world without introducing depth bias into the scene representations.

\subsection{Hierarchical Semantic Guidance} \label{sec:HSG}
Inspired by the diverse variations in semantic consistency on multi-resolution images, we propose hierarchical semantic guidance to supervise the coarse-to-fine semantic content corresponding to the coarse-to-fine scene representations with the CLIP encoder.

\begin{figure}[!t]
\centering
\includegraphics[width=.95\linewidth]{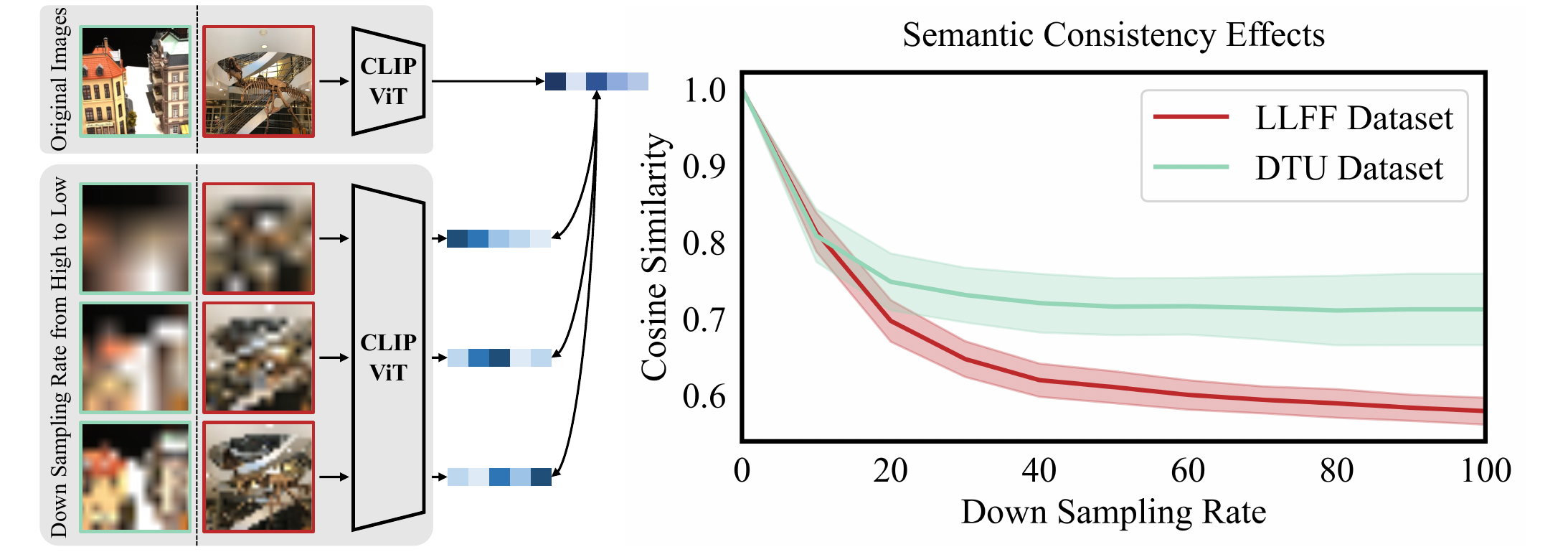}
\caption{\textbf{Semantic Consistency Effects on Multi-Resolution Images.} By computing the cosine similarity of the feature vectors between the original image and its down sampling results, we can find that the cosine similarity decreases sharply with the increase of down sampling rate. Especially when the dataset is contributed by complex real-world scenes~(e.g., LLFF dataset~\cite{mildenhall2019local}), meaning that an image contains plenty of content, the effect on semantic consistency supervision is further limited.}
\label{fig:hsg}
\vspace{-10pt}
\end{figure}

As shown in Fig.~\ref{fig:hsg}, the effect of CLIP is limited for supervising the semantic consistency over multi-resolution images. NeRF first learn low-frequency information for coarse scene representations~\cite{lin2021barf}, resulting in images rendered at the beginning of training that resemble those low-resolution images containing little content. Therefore, it is difficult to match the feature vectors encoded from these rendered images with those encoded from high-resolution original images.
 
To solve this problem, the proposed HSG randomly selects a known camera pose and performs the semantic supervision between the rendered images and the original images of the same scale by sampling a set of pixels~(rays) $\mathcal{P}$ using a coarse-to-fine grid sampling strategy, which can be formulated as:
\begin{equation}
\mathcal{P}(s_{i})
=
\left\{\left(u, v\right)
\mid 
{u}\in{\left[0,H,stride=s_{i}\right]},
{v}\in{\left[0,W,stride=s_{i}\right]}
\right\}
\end{equation}
where $(u,v)$ denotes pixel location and $s_{i}$ denotes sampling stride, which can be formulated as:
\begin{equation}
s_{i} = \max \left( ceil\left( 
s_{max} \cdot \frac{1+\cos\left({\left(i/N_{hsg}\right)}\pi\right)}{2}
\right), 1\right)
\end{equation} 
Note that $s_{max}$ denotes the maximum sampling stride, $N_{hsg}$ denotes how many iterations until the full pixels are sampled, and $ceil$ denotes rounding operation. Then, feature vectors of the sampled $\mathbf{\hat{C}}_{sem}$ and $\mathbf{\hat{C}}_{i}$ are encoded by CLIP $f_{clip}$, which can be formulated as:
\begin{equation}
    \varphi_{sem} = f_{clip}\left( \mathbf{\hat{C}}_{sem} \right)
    \text{, }
    \varphi_{i} = f_{clip}\left( \mathbf{\hat{C}}_{i} \right)
\end{equation}
The coarse-to-fine semantic consistency supervision thus can be performed, since more scene content is aggregated into the feature vectors with the sampling stride $s_{i}$ gradually decreasing to 1.

\subsection{Reconstruction Loss Function}
As we discuss in Sec.~(\ref{sec:HGG}), HG$^{3}$-NeRF represent the neural radiance fields without using explicit depth supervision for sparse view inputs. Therefore, our reconstruction loss is constructed with hierarchical photometric guidance and hierarchical semantic guidance.

\textbf{HPG Loss Function.} We follow the loss function $\mathcal{L}_{hpg}$~(Eq.~(\ref{eq:hpg})) proposed in NeRF to supervise the reconstruction of the scene appearance.

\textbf{HSG Loss Function.} The HSG method allows HG$^{3}$-NeRF to supervise the the coarse-to-fine semantic consistency by computing the cosine similarity $\mathcal{L}_{hsg}$ between $\varphi_{sem}$ and $\varphi_{i}$, which can be formulated as: 
\begin{equation}
    \mathcal{L}_{hsg} = \varphi_{sem}^{\mathrm{T}} \cdot \varphi_{i}
\end{equation}
Finally, we use these two loss functions to optimize the scene representations, which can be formulated as the following minimization problem:
\begin{equation}
\min_{{\Theta_c},{\Theta_f}}
\left(
\mathcal{L}_{hpg} + \lambda \mathcal{L}_{hsg}
\right)
\end{equation}
where $\lambda$ denotes the weighting factor for $\mathcal{L}_{hsg}$.

\section{Experiments}
\textbf{Dataset.} We validate the proposed HG$^{3}$-NeRF using the standard datasets of different levels: DTU dataset~\cite{jensen2014large} and LLFF dataset~\cite{mildenhall2019local}. DTU dataset consists of object-level scenes where the image content contains objects on a white table with a black background. LLFF dataset consists of real-world-level scenes with sequentially captured images from hand-held cameras.

\textbf{Baselines.} We compare against state-of-the-art baselines including pre-training methods and per-scene optimization methods. Moreover, we adopt \textit{'w sdp'} to label the method using the sparse depth prior and \textit{'ft'} to label the method to fine-tune for each scene. Note that $^\dagger$DS-NeRF is our re-implementation, since there is no direct way to evaluate DS-NeRF in their released code.

\textbf{Evaluation Metrics.} We follow the official evaluation metrics and thus adopt the mean of PSNR, structural similarity index~(SSIM)~\cite{wang2004image}, and the LPIPS~\cite{zhang2018unreasonable} perceptual metric.

\subsection{Implement Details}
We implement the framework of HG$^3$-NeRF based on NeRF. The pre-training baselines are pre-trained on the large-scale DTU dataset. The regularization baselines are trained from scratch for each scene. We use the Adam optimizer with a learning rate of $1\times10^{-3}$ exponentially decaying to $1\times10^{-5}$ and train our method on two NVIDIA V100 GPUs. We sample $64$ volume points for the coarse model and $128$ volume points for the fine model. We set $N_{hgg}$ and $N_{hsg}$ as $10\%$ and $50\%$ of total training iterations. We set $s_{max}=0.1\cdot\min (H,W)$, $\epsilon_{hgg}=0.2$, and $\lambda=0.2$ where $H,W$ denote the height and width of the image. Following experimental protocols of baselines~\cite{niemeyer2022regnerf,yu2021pixelnerf}, we compare the performance of HG$^{3}$-NeRF against these baselines for the scenarios of 3,6, and 9 views.

\begin{figure}[!t]
\centering
\includegraphics[width=.9\linewidth]{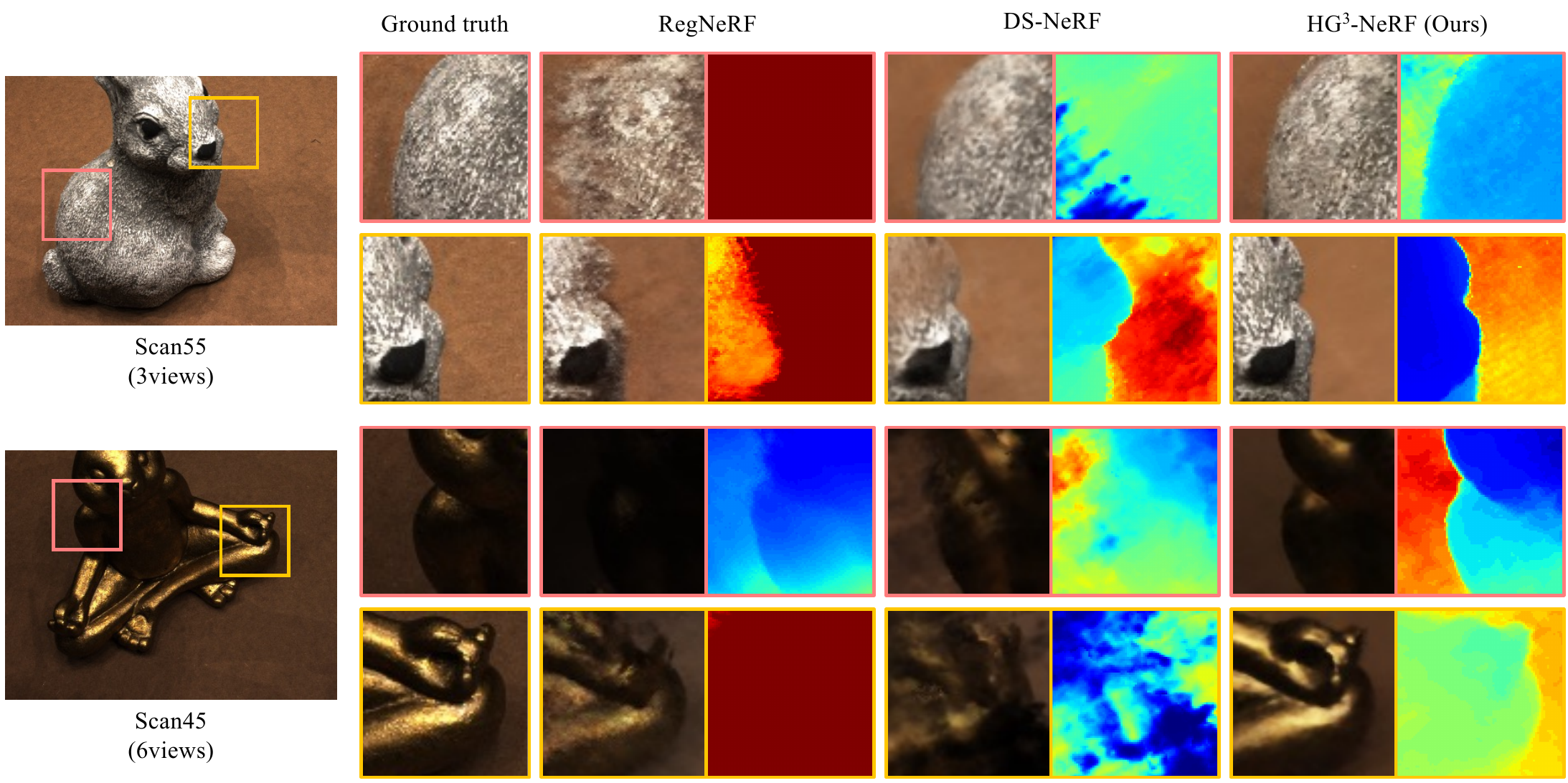}
\caption{\textbf{Qualitative Results on DTU Dataset.} HG$^3$-NeRF maintain more fine details in the synthetic images and more sharper edges in the estimated depth maps.}
\label{fig:dtu}
\vspace{-10pt}
\end{figure}

\begin{figure}[!t]
\centering
\includegraphics[width=.9\linewidth]{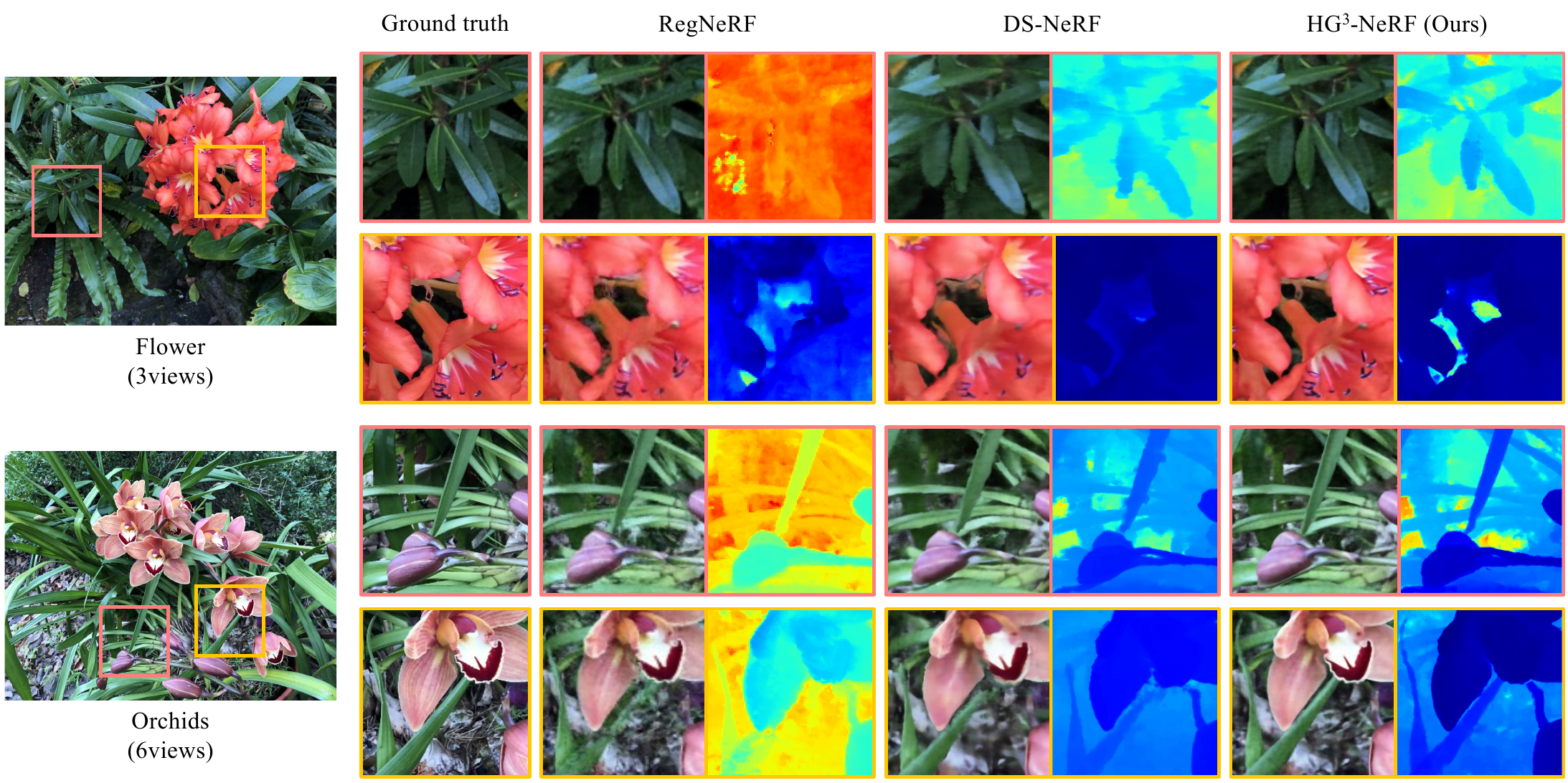}
\caption{\textbf{Qualitative Results on LLFF Dataset.} Our method synthesizes high-fidelity images and estimates accurate depth maps in the forward-facing scene without NDC space.}
\label{fig:llff}
\vspace{-10pt}
\end{figure}

\subsection{Comparison with SOTA Baselines}
\textbf{Comparison on DTU Dataset.} We first evaluate the performance of our method in the object-level DTU dataset. Fig.~\ref{fig:dtu} presents the qualitative results on object-level DTU dataset. HG$^{3}$-NeRF can synthesize images with more fine details and estimate depth maps with distinct edges in comparison to other SOTA baselines. As we discuss in Sec.~\ref{sec:HGG}, bias can be introduced into the scene representations through the loss supervision function contributed by the depth prior. Our HG$^{3}$-NeRF incorporate geometric consistency using the HGG method, where the sparse depth prior is employed to guide a local-to-global volume sampling rather than a loss function. Additionally, our HSG method can further improve the semantic consistency of the scene representations by computing a coarse-to-fine cosine similarity. Therefore, as illustrated in Table.~\ref{tab:dtu}, HG$^{3}$-NeRF can outperform most SOTA baselines with high consistency between seen and unseen views even with only 3 views available~(PSNR=19.37). Moreover, as the semantic content is aggregated into the scene representation from coarse to fine, the synthesis quality can be further improved~(PSNR=25.87, 9 views) in comparison to DietNeRF~(23.83, 9 views), which directly compute the cosine similarity between rendered images and original images.

\textbf{Comparison on LLFF Dataset.} 
We then evaluate HG$^3$-NeRF on complex real-world LLFF dataset. 
As the qualitative results presented in Fig.~\ref{fig:llff}, different from some baselines~(e.g. RegNeRF) that require the NDC space to represent the forward-facing scene, our method can realize high-fidelity view synthesis under the real-world space. Moreover, as shown in Table.~\ref{tab:llff}, the effect of semantic consistency for complex scenes is severely limited~(PSNR=14.94, DietNeRF for 3 views). The HSG method can utilize the coarse-to-fine semantic consistency to gradually aggregate the semantic content into the representations. As a result, our HG$^{3}$-NeRF is able to employ the hierarchical geometric, semantic, and photometric guidance to represent the complex scene with high consistency in appearance, geometry, and semantics even from sparse view inputs~(PSNR=20.98, 3 views).

\begin{table*}[!ht]
\centering
\caption{\textbf{Quantitative Results on DTU Dataset.} The best is in bold. Conducted with sparse view inputs, the proposed HG$^3$-NeRF can achieve better view synthesis quality than most SOTA baselines.}
\setlength\tabcolsep{12pt}
\resizebox{.95\linewidth}{!}{
\begin{tabular}{l|c|ccc|ccc|ccc}
\toprule
& \multirow{2}{*}{Configures} 
& \multicolumn{3}{c|}{PSNR $\uparrow$} 
& \multicolumn{3}{c|}{SSIM $\uparrow$} 
& \multicolumn{3}{c}{LPIPS $\downarrow$} \\
&  
& 3-view & 6-view & 9-view  
& 3-view & 6-view & 9-view 
& 3-view & 6-view & 9-view \\
\midrule\midrule
PixelNeRF & \multirow{3}{*}{Pre-training} 
& 16.82 & 19.11 & 20.40 
& 0.695 & 0.745 & 0.768 
& 0.270 & 0.232 & 0.220 \\
SRF & 
& 15.32 & 17.54 & 18.35 
& 0.671 & 0.730 & 0.752 
& 0.304 & 0.250 & 0.232 \\
MVSNeRF &  
& 18.63 & \cellcolor{orange!25}{20.70} & 22.40 
& \cellcolor{red!25}\textbf{0.769} & \cellcolor{yellow!25}{0.823} & \cellcolor{yellow!25}{0.853}
& \cellcolor{yellow!25}{0.197} & 0.156 & \cellcolor{yellow!25}{0.135} \\
\midrule
PixelNeRF \textit{ft} & \multirow{3}{*}{\shortstack{Pre-training\\and\\Fine-tune}}
& \cellcolor{orange!25}{18.95} & 20.56 & 21.83 
& 0.710 & 0.753 & 0.781 
& 0.269 & 0.223 & 0.203 \\
SRF \textit{ft} & 
& 15.68 & 18.87 & 20.75 
& 0.698 & 0.757 & 0.785 
& 0.281 & 0.225 & 0.205 \\
MVSNeRF \textit{ft} & 
& 18.54 & 20.49 & 22.22 
& \cellcolor{red!25}\textbf{0.769} & 0.822 & \cellcolor{yellow!25}{0.853} 
& \cellcolor{yellow!25}{0.197} & \cellcolor{yellow!25}{0.155} & \cellcolor{yellow!25}{0.135} \\
\midrule
DietNeRF & \multirow{4}{*}{\shortstack{Per-scene\\\\Optimization}} 
& 11.85 & 20.63 & \cellcolor{yellow!25}{23.83} 
& 0.633 & 0.778 & 0.823 
& 0.314 & 0.201 & 0.173 \\
RegNeRF & 
& \cellcolor{yellow!25}{18.89} & \cellcolor{yellow!25}{22.20} & \cellcolor{orange!25}{24.93}
& \cellcolor{yellow!25}{0.745} & \cellcolor{orange!25}{0.841} & \cellcolor{orange!25}{0.884}
& \cellcolor{orange!25}{0.190} & \cellcolor{orange!25}{0.117} & \cellcolor{orange!25}{0.089} \\
$^\dagger$DS-NeRF \textit{w sdp} & 
& {16.29} & {19.07} & {21.98}
& {0.559} & {0.807} & {0.828} 
& {0.451} & {0.211} & {0.195} \\
\textbf{HG$^{3}$-NeRF} \textit{w sdp} & 
& \cellcolor{red!25}{\textbf{19.37}} & \cellcolor{red!25}{\textbf{23.35}} & \cellcolor{red!25}{\textbf{25.87}}
& \cellcolor{orange!25}{0.759} & \cellcolor{red!25}{\textbf{0.855}} & \cellcolor{red!25}{\textbf{0.891}}
& \cellcolor{red!25}{\textbf{0.177}} & \cellcolor{red!25}\textbf{0.094} & \cellcolor{red!25}\textbf{0.061} \\
\bottomrule
\end{tabular}}
\label{tab:dtu}
\vspace{-10pt}
\end{table*}

\begin{table*}[!ht]
\centering
\caption{\textbf{Quantitative Results on LLFF Dataset.} The best is in bold. For the complex real-world scenes, HG$^{3}$-NeRF can outperform other SOTA baselines without using NDC space.}
\setlength\tabcolsep{12pt}
\resizebox{.95\linewidth}{!}{
\begin{tabular}{l|c|ccc|ccc|ccc}
\toprule
& \multirow{2}{*}{Configures} 
& \multicolumn{3}{c|}{PSNR $\uparrow$} 
& \multicolumn{3}{c|}{SSIM $\uparrow$} 
& \multicolumn{3}{c}{LPIPS $\downarrow$} \\
&  
& 3-view & 6-view & 9-view  
& 3-view & 6-view & 9-view 
& 3-view & 6-view & 9-view \\
\midrule\midrule
PixelNeRF & \multirow{3}{*}{Pre-training}
& 7.93  & 8.74  & 8.61 
& 0.272 & 0.280 & 0.274 
& 0.682 & 0.676 & 0.665 \\
SRF &
& 12.34 & 13.10 & 13.00 
& 0.250 & 0.293 & 0.297 
& 0.591 & 0.594 & 0.605 \\
MVSNeRF & 
& 17.25 & 19.79 & 20.47 
& 0.557 & 0.656 & 0.689 
& 0.356 & 0.269 & 0.242 \\
\midrule 
PixelNeRF \textit{ft} & \multirow{3}{*}{\shortstack{Pre-training\\and\\Fine-tune}}
& 16.17 & 17.03 & 18.92 
& 0.438 & 0.473 & 0.535 
& 0.512 & 0.477 & 0.430 \\
SRF \textit{ft} &
& 17.07 & 16.75 & 17.39 
& 0.436 & 0.438 & 0.465 
& 0.529 & 0.521 & 0.503 \\
MVSNeRF \textit{ft} & 
& 17.88 & 19.99 & 20.47 
& 0.584 & 0.660 & 0.695 
& \cellcolor{orange!25}{0.327} & 0.264 & 0.244 \\
\midrule 
DietNeRF & \multirow{4}{*}{\shortstack{Per-scene\\\\Optimization}} 
& 14.94 & 21.75 & \cellcolor{yellow!25}{24.28} 
& 0.370 & \cellcolor{yellow!25}{0.717} & \cellcolor{yellow!25}{0.801} 
& 0.496 & \cellcolor{yellow!25}{0.248} & \cellcolor{yellow!25}{0.183} \\
RegNeRF & 
& \cellcolor{yellow!25}{19.08} & \cellcolor{orange!25}{23.10} & \cellcolor{orange!25}{24.86} 
& \cellcolor{yellow!25}{0.587} & \cellcolor{orange!25}{0.760} & \cellcolor{orange!25}{0.820} 
& \cellcolor{yellow!25}{0.336} & \cellcolor{orange!25}{0.206} & \cellcolor{orange!25}{0.161} \\
$^\dagger$DS-NeRF \textit{w sdp} & 
& \cellcolor{orange!25}{19.68} & \cellcolor{yellow!25}{22.45} & {23.78}
& \cellcolor{orange!25}{0.615} & {0.674} & {0.722}
& {0.403} & {0.356} & {0.324} \\
\textbf{HG$^3$-NeRF} \textit{w sdp} & 
& \cellcolor{red!25}{\textbf{20.98}} & \cellcolor{red!25}{\textbf{24.60}} & \cellcolor{red!25}{\textbf{25.51}}
& \cellcolor{red!25}{\textbf{0.682}} & \cellcolor{red!25}{\textbf{0.823}} & \cellcolor{red!25}{\textbf{0.844}}
& \cellcolor{red!25}{\textbf{0.198}} & \cellcolor{red!25}{\textbf{0.107}} & \cellcolor{red!25}{\textbf{0.075}} \\
\bottomrule
\end{tabular}}
\label{tab:llff}
\vspace{-10pt}
\end{table*}

\begin{figure}[!ht]
\centering
\begin{minipage}{.45\linewidth}
\includegraphics[width=.9\linewidth]{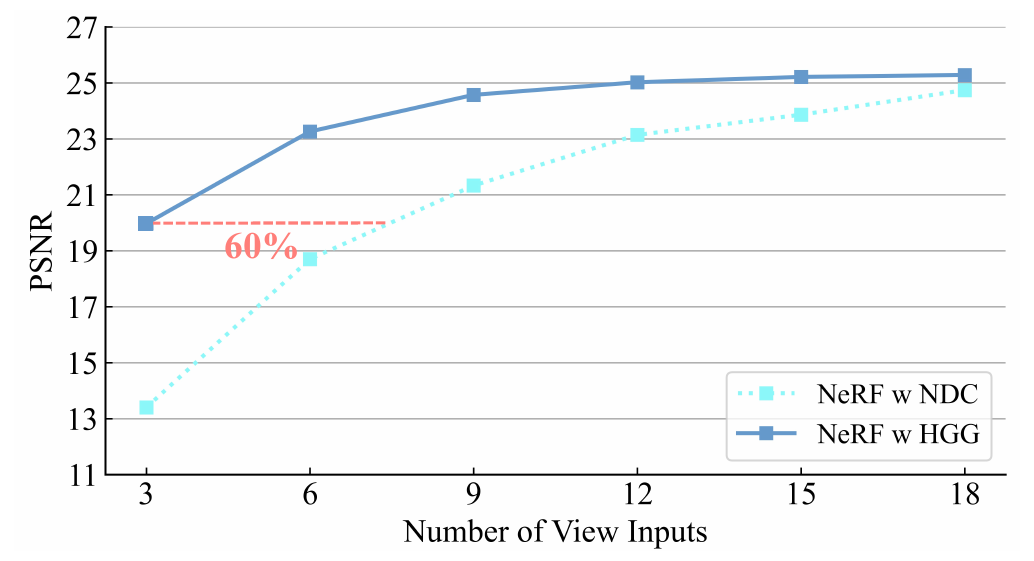}
\caption{\textbf{Data and Space Efficiency.} 
In sparse settings, NeRF with HGG~(in real-world space) require up to 60$\%$ fewer images than NeRF using NDC space to achieve comparable performance on LLFF dataset.}
\label{fig:com}
\end{minipage}
\quad
\begin{minipage}{.45\linewidth}
\centering
\captionof{table}{\textbf{Ablation Study.} We investigate the effect of HGG and HSG on LLFF dataset. For sparse view inputs, HGG serves the crucial role of representations in the real-world space and avoiding overfitting problems. HSG can further improve the performance with the coarse-to-fine semantic guidance.}
\setlength\tabcolsep{8pt}
\resizebox{.9\linewidth}{!}{
\begin{tabular}{cc|ccc}
\toprule
\multicolumn{2}{c|}{Configures}
& \multicolumn{3}{c}{PSNR} \\
HGG & HSG 
& 3-view & 6-view & 9-view \\
\midrule\midrule
\checkmark &  & \cellcolor{orange!25}{19.97} & \cellcolor{orange!25}{23.46} & \cellcolor{orange!25}{24.57} \\
& \checkmark & \cellcolor{yellow!25}{15.82} & \cellcolor{yellow!25}{22.71} & \cellcolor{yellow!25}{23.29} \\
\checkmark & \checkmark & \cellcolor{red!25}\textbf{20.98} & \cellcolor{red!25}\textbf{24.06} & \cellcolor{red!25}\textbf{25.51} \\
\bottomrule
\end{tabular}}
\label{tab:abs}
\end{minipage}
\vspace{-10pt}
\end{figure}

\subsection{Model Analysis}
\textbf{Ablation Study.}
We conduct an ablation study to investigate the effect of HGG and HSG under 3,6,9 view inputs on LLFF dataset. As shown in Table.~\ref{tab:abs}, since we represent the scene in the real-world space, the HGG method can maintain the most important geometric consistency for different viewpoints and thus plays a key role in our method, especially for only 3 view inputs. Moreover, the HSG method can supervise the semantic consistency for the coarse-to-fine representations to further improve the synthesis quality based on the HGG method. These two methods can be regarded as the convenient tools for improving the performance of novel view synthesis.

\textbf{NDC space vs Real-World Space.}
We evaluate the data and space efficiency by comparing the performance of NeRF with HGG~(real-world space) and NeRF with NDC space. As shown in Fig.~\ref{fig:com}, for sparse view inputs, NeRF with HGG only requires up to 60$\%$ fewer view inputs to achieve the view synthesis quality that is comparable to NeRF with NDC space on the test image set. Therefore, in addition to improving the performance under sparse view inputs, the proposed HGG method can also replace the NDC space for representing the forward-facing scenes.

\section{Conclusion}
In this paper, the proposed HG$^{3}$-NeRF exploit the hierarchical geometric, semantic, and photometric guidance to represent the scene from sparse views. We propose HGG to leverage the sparse depth prior to indicate a local-to-global region for volume sampling without introducing geometric bias. The HSG method maintains the semantic consistency by computing the coarse-to-fine cosine similarly. The experimental results demonstrate that the proposed HG$^{3}$-NeRF can synthesize high-fidelity images and estimate accurate depth maps with distinct edges.

\textbf{Limitations.} In our method, the camera poses are still required to be estimated from SfM. Note that sparse view inputs can also affect the accuracy of pose estimation, and some works prove that noisy camera poses can degrade the view synthesis quality~\cite{lin2021barf,chng2022gaussian}. Therefore, there is still scope to improve the performance of NeRF from sparse view inputs. In the future work, we will follow the methods proposed in this paper and extend them on solving the joint optimization problem about camera poses and NeRF from sparse view inputs.

\bibliographystyle{plain}
\bibliography{neurips_2023}


\end{document}